# Traffic Sign Recognition Dataset and Data Augmentation

Jingzhan Ge[b, †]

School of Control Science and Engineering, Tiangong University

Tianjin, 300387, China

[b]1910410157@tiangong.edu.cn

**ABSTRACT**

**Although there are many datasets for traffic sign classification, there are few datasets collected for traffic sign recognition and few of them obtain enough instances especially for training a model with the deep learning method. The deep learning method is almost the only way to train a model for real-world usage that covers various highly similar classes compared with the traditional way such as through color, shape, etc. Plus, due to the appearance frequency of different classes of traffic signs in the real world, the imbalance between different classes' instances in the datasets makes the training results even worse. Also, for some certain sign classes, their sign meanings were destined to can't get enough instances in the dataset. To solve this problem, we purpose a unique data augmentation method for the traffic sign recognition dataset that takes advantage of the standard of the traffic sign. We called it TSR dataset augmentation. We based on the benchmark Tsinghua-Tencent 100K (TT100K) dataset to verify the unique data augmentation method. we performed the method on four main iteration version datasets based on the TT100K dataset and the experimental results showed our method is efficacious. The iteration version datasets based on TT100K, data augmentation method source code and the training results introduced in this paper are publicly available.**

## 1. INTRODUCTION

Deep learning is a machine learning technique that teaches computers to do what humans are born with: learn by example. In deep learning, computer models learn to perform tasks directly from images, text, or sound. Deep learning models can achieve state-of-the-art precision, sometimes exceeding human levels. Models are trained by using a large set of labeled data[1]. The dataset is vitally important for deep learning. The first thing we do was to choose a suitable traffic sign recognition dataset. For well-known TSR datasets like DFG Traffic Sign Dataset, Swedish Traffic Signs Dataset, German Traffic Sign Detection Benchmark, and LISA Traffic Sign Dataset are collected outside mainland China. Chinese TRS datasets are much larger: CCTS TSDD, CCTSDB and Tsinghua-Tencent 100K (TT100K). However, these datasets have the following shortcomings in varying degrees, such as excessive subdivision or crude classification classes, not enough instances, and dataset collection methods. We think that the best dataset collection method is collected by the camera. For example, if collected through video recording, one picture is collected every 5 frames in the video clips containing traffic signs. This method will cause a large number of similar scenes, especially in the case of the small size of the dataset, which is unfavorable for model training and for our data augmentation method for traffic sign recognition datasets purposed in this paper.

DFG Traffic Sign Dataset was collected in Slovenia and labeled by DFG Consulting d.o.o. It contains 6957 images and 200 classes, each containing at least 20 instances. Each class containing at least 20 instances is obviously not enough for training[2]. The German Traffic Sign Detection Benchmark features a single-image detection problem with 900 images (divided into 600 training images and 300 evaluation images) and a total of 1206 instances divided into only 4 classes. Apparently, this dataset's class is too crude[3]. The Swedish Traffic Signs Dataset was collected in Sweden. It contains

20,000 images but only 20% of them (4000 images) have been labeled into 6 classes. The total instance is only 3488, which is too small. Also, it has another disadvantage is that the dataset is collected through video recording[4]. LISA Traffic Sign Dataset is collected by the Intelligent Safety Driverless Lab of the University of California, San Diego. It also collected images through video capture frames, including 47 classes of US traffic signs with a total of 6610 images and 7855 instances[5].

Compared with foreign TSR datasets, Chinese TSR datasets are much larger, contain more pictures and instances and are collected by cameras, especially by street view image acquisition, which is better for model training. Chinese Traffic Sign Database TSDD was collected by Beijing Jiaotong University. The TSDD includes 10000 traffic scene images containing 58 classes. The dataset is collected by the camera under nature scenes or from BAIDU Street View. CCTSDB 2021 was collected by the Changsha University of Science and Technology. This dataset is based on the Chinese Traffic Sign Database TSDD dataset[6]. There are nearly 20,000 images in total, but they are divided into only three categories, which are too crude: indication signs, prohibition signs and warning signs. The Tsinghua-Tencent 100K 2021 dataset is published by the joint laboratory of Tsinghua University and Tencent[7]. The dataset consists of 100,000 images collected from Tencent Street View, among which 9,170 images contain traffic signs with a total of 27,346 instances divided into 201 different classes.

Table 1. Comparison of TSR Datasets

|  | Picture with Sign | Number of Class | Type |
| --- | --- | --- | --- |
| **DFG TSD** | 6957 | 200 | Camera |
| **Swedish TSD** | 4000 | 6 | Video Frame |
| **GTSDB** | 900 | 4 | Camera |
| **LISA TSD** | 6610 | 47 | Video Frame |
| **CCTS TSDD** | 10000 | 45 | Camera Shoot |
| **CCTSDB 2021** | 17856 | 3 | Camera Shoot |
| **TT100K 2021** | 9170 | 201 | Camera Shoot |

The above table is the comparison of these 7 TSR datasets. Considering the number of classes, dataset collecting methods, total instances, publish time and other factors, we decide to use Tsinghua-Tencent 100K (TT100K), Which has about 10,000 images and the 201 classes allows us to verify our unique data augmentation method for traffic sign recognition dataset.

## 2. DATA AUGMENTATION

The performance of most deep learning models depends on the quality, quantity and relevancy of training data. However, insufficient data is one of the most common challenges in implementing machine learning in the enterprise. This is because collecting such data can be costly and time-consuming in many cases. Data augmentation is a set of techniques to artificially increase the amount of data by generating new data points from existing data. This includes making small changes to data or using deep learning models to generate new data points. For image classification, detection and segmentation data augmentation, the common ways are padding, random rotating, re-scaling, vertical and horizontal flipping, translation (image is moved along X, Y direction), cropping, zooming, darkening & brightening/color modification, gray scaling, changing contrast, adding noise, random erasing[8].

In the TSR training process, we do need to use some common data augmentation ways. However, some of the common ways can't be used in TSR. For example, the right turn and left turn signs, if we use rotating or flipping, will obfuscate the signs. We used the YOLOv5's mosaic augmentation. The mosaic applies online image space and color space augmentations in the training loader to present a new and unique augmented mosaic (original image + 3 random images) each time an image is loaded for training.

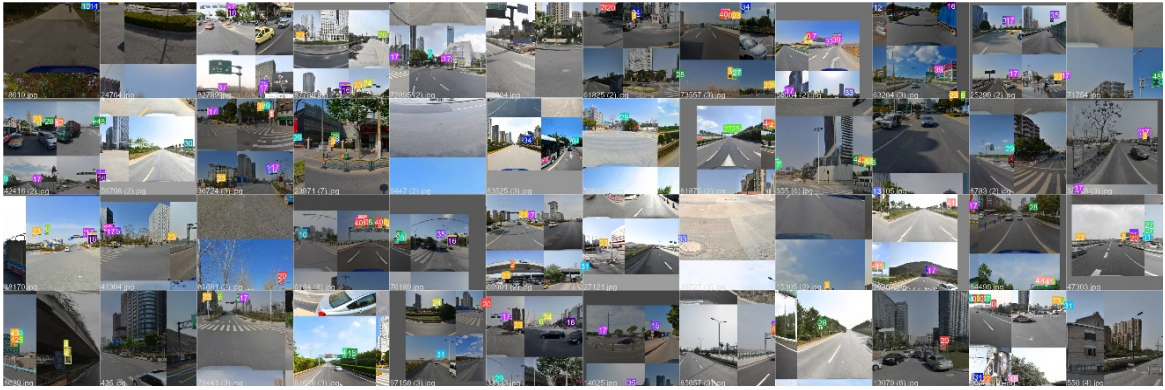

Figure 1. Training batch with mosaic augmentation

These conventional ways can only improve the generalization ability of the model but do not address problems such as some classes' lack of a large number of instances in the TSR dataset and the extreme imbalance in the number of class instances. So, we purpose a unique data augmentation method for the traffic sign recognition dataset that takes advantage of the standard of the traffic sign. We called it TSR dataset augmentation.

The TSR dataset augmentation aims to solve the problem of some classes' lack of instances and imbalance in the number of class instances. It takes advantage of the standard of the traffic sign. The main idea of TSR dataset augmentation is that with the coordinates, and class numbers in the annotation file, we can replace the images' labeled section with other class's marks. For example, the p10 class in the dataset has 374 instances, we use some of the images in the dataset and replace their labeled sections with p10's mark, we can easily get more p10 instances. This method could work, mostly because the traffic signs are 2D objects, which is different from other conventional object detection tasks which are 3D objects. while for 3D objects, for example, in the COCO dataset the class of car, the differences in colors, shapes, sizes and angles. Even if the front view and the side view of the same car are different, we can't use a mark to replace the labeled section[9]. But for the TSR dataset, we can do that. For 2D objects, the difference of the target in the image is only in the angle, brightness and shade of the light (back-lighting, front-lighting), and according to the Road Traffic Signs and Markings Part 2 – Road Traffic Signs, we can get the shape, size scale, and meaning of the traffic sign[10]. Each class's signs are exactly the same in color and shape. We found that most of the signs' shapes are round which means most signs are in the same shape, we can replace each sign with the same shape. And some signs in the images taken at a specific angle can also be scaled and adjusted according to the annotation files' coordinates. For brightness and shade of the light, in order to make the augmentation images close to reality, the brightness of the replaced marks processed random brightness to ensure the diversity of data.

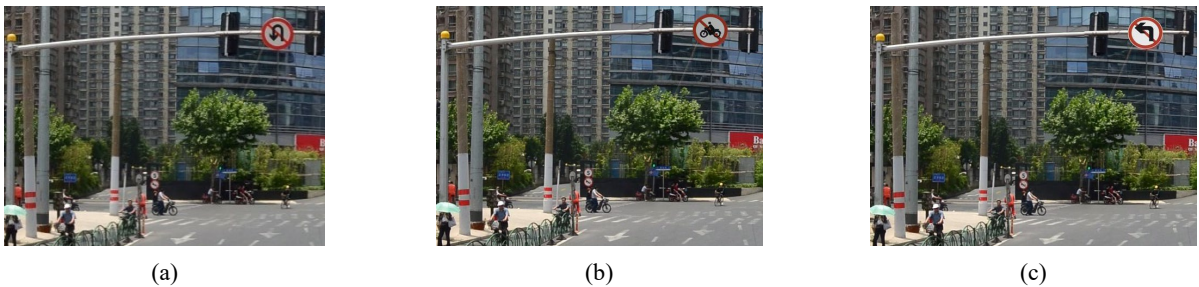

(a)          (b)          (c)

Figure 2. (A)193.jpg Original image; (B), (C)TSR dataset augmentation p12&p23

The TSR dataset augmentation's main steps are: First, read the coordinates and class number in the annotation file. Second, paste the mark of the class to be augmented into the annotation coordinate area. Third, rewrite the annotation file. This is highly dependent on the accuracy of annotation, and the annotation of the TT100K dataset is very accurate. Except for a few annotations due to the deformation of the panoramic camera, the majority of data images can be augmented by this method.

The YOLO annotation format is saved in a .txt file. A total of 5 pieces of data are separated by spaces. Through these data, the marks can be scaled and adjusted to restore the specific angle on the images.

```
17         0.436767578125 0.2734375  0.05615234375 0.05078125
Class_id; x;              y;         w;            h
```

The first data Class_id is the class number. The second data x is the ratio of center abscissa to image width. The third data y is the ratio of center ordinate to image height. The fourth data w is the ratio of bounding box width to image width. The fifth data h is the ratio of bounding box height to image height. The steps for calculating coordinates for augmentation are as follows.

First, do bounding box inverse normalization. Assuming that the height and width of the image are $h1$ and $w1$ respectively.

$$x\_t = x * w1$$
$$y\_t = y * h1$$
$$w\_t = w * w1$$
$$h\_t = h * h1$$

Second, calculate coordinates. The coordinate of the top left corner of the bounding box is $(top\_left\_x, top\_left\_y)$, and the coordinate of the bottom right corner is $(bottom\_right\_x, bottom\_right\_y)$ can be obtained as:

$$top\_left\_x = x\_t - w\_t / 2$$
$$top\_left\_y = y\_t - h\_t / 2$$
$$bottom\_right\_x = x\_t + w\_t / 2$$
$$bottom\_right\_y = y\_t + h\_t / 2$$

Third, get the width ($icon\_w$) and height ($icon\_h$) of the mark that will be pasted on the labeled section.

$$icon\_w = bottom\_right\_x - top\_left\_x$$
$$icon\_h = bottom\_right\_y - top\_left\_y$$

Through this method, it can realize the augmentation of specific classes so that the Insufficient instances and imbalance between various classes can be reduced. And can achieve the purpose of using the collected limited real-world data, after augmentation, using the least amount of data, to train the model containing complex classes. The details of the TSR dataset augmentation steps will be introduced in the latter section.

## 3. DATASET

The TT100K images are derived from Tencent Street View panoramas taken by six high-pixel wide-angle SLRS in various cities in China, with varying lighting and weather conditions. The resolution of the original Street View panorama was 8192x2048, and then the panorama was cut into four parts, and the size of the images in the dataset was 2048x2048.

In TT100K, a total of 201 different classes appeared. The number of instances contained by each class is shown in figure 3 below: Among the 201 class, 84 classes have less than 10 instances; Class 62 is 10-75 instances; Only 45 classes have more than 100 instances.

In deep learning, the instances of each class need to be sufficient. The more instances you have, the better effect of the trained model and the stronger the generalization ability of the model. However, according to Ultralytics's suggestion for training custom datasets using YOLOv5, each class's image should be larger than 1500, and the number of instances of each class should be larger than 10000. The number of instances of each type in the TT100K dataset is far insufficient.

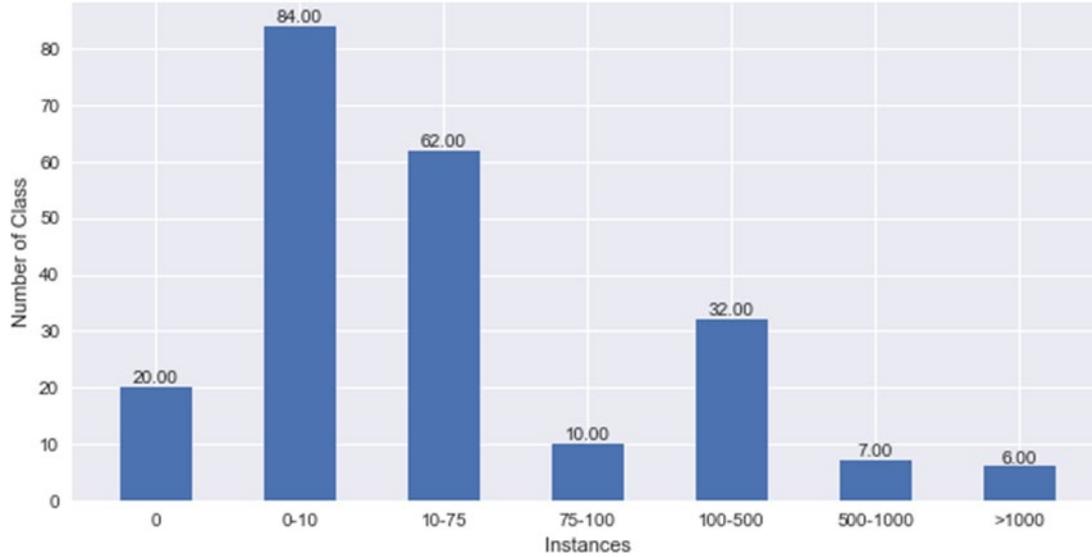

Figure 3. TT100K instances per class

Based on the TT100K dataset, this paper iterates 4 versions of the dataset to validate TSR dataset augmentation. Version 1 is used as a baseline dataset to compare and validate our TSR dataset augmentation. Version 2 is based on Version 1 and merges some classes into one class. It is the first time to use TSR dataset augmentation and we get the initial validation. Version 3 is based on the benchmark TT100K dataset and follows the idea of Version 2 with more TSR dataset augmentation. Version 4 is also based on the benchmark TT100K dataset, it doubles the images and triples the instances, it is the final version and completely validates the effectiveness of TSR dataset augmentation.

## 3.1 Version 1 Dataset

In deep learning, too few instances of each class will lead to bad training results, so it is necessary to ensure that the number of instances of each class is enough. In view of a large number of classes in the TT100K dataset with a few instances, Version 1.0 adopts the simplest and naive idea that only selects classes with more than 100 instances and there are 45 classes. As shown in the figure below, orange is the average number of instances = 515. As only 45 classes were left, the total number of instances was reduced from 27,346 to 23,182.

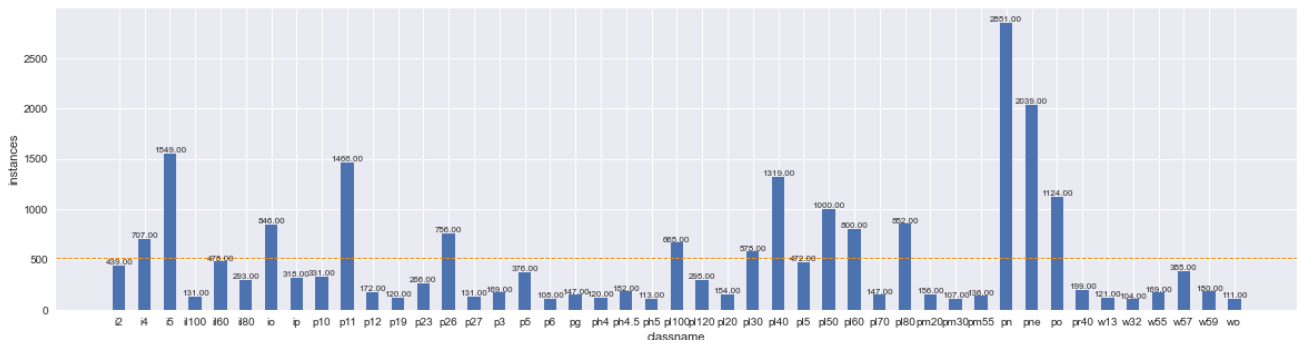

Figure 4. Version 1 Dataset instances per class

We used the YOLOv5s Baseline model for training, and the training results were shown in the following table. mAP@.5:.95 only reached 0.6546.

**Table 2. Version 1 Dataset Training Results**

| Dataset | Class | Picture | Instances | mAP@.5 | Improve | mAP@.5:.95 | Improve |
|---|---|---|---|---|---|---|---|
| Version1 | 45 | 9170 | 23182 | 0.8518 | / | 0.6546 | / |

These 45 classes are only part of the extracted classes. The effect of the model's TSR in the real environment is very poor.

Such as speed limit signs. Version 1.0's class doesn't include pl90 (Speed Limit 90), and when pl90 is encountered in the real environment test, the sign would be misidentified as another Speed Limit class. This results in models that have no practical value at all. Given that the mAP of the Version 1.0 dataset is not very good, we think it is because of instance insufficient of many classes, especially in similar classes such as the subclasses of pl (Maximum Speed limit) like pl120 and pl20. Therefore, we think that these subclasses can be merged into one class to offset the poor training results caused by the instance insufficient. The Version 2 dataset will validate this conjecture and preliminarily validate the TSR dataset augmentation proposed in this paper.

### 3.2  Version 2 Dataset

The Version 2 dataset is available in two versions, 2.0 and 2.1. Version 2.0 is based on Version 1, without data augmentation, it merged some classes into one class, such as pl subclasses (all levels of speed limit sign) into a single class pl (speed limit) to solve the instance insufficiently. And make the training model based on the limited dataset, having certain practical application abilities. The figure below shows how the specific classes are merged. In this way, the dataset was reduced from 45 classes to 24 classes.

| pl100 | pl120 | pl20 | pl30 | pl40 | pl5 | pl50 | pl60 | pl70 | pl80 | p19 | p23 |
|-------|-------|------|------|------|-----|------|------|------|------|-----|-----|
| pl | | | | | | | | | | pt | |
| w13 | w32 | w55 | w57 | w59 | wo | pm20 | pm30 | pm55 | ph4 | ph4.5 | ph5 |
| w | | | | | | pm | | | ph | | |

Figure 4. Merged Classes

The Version 2.1 Dataset Instances per Class are shown in the figure below

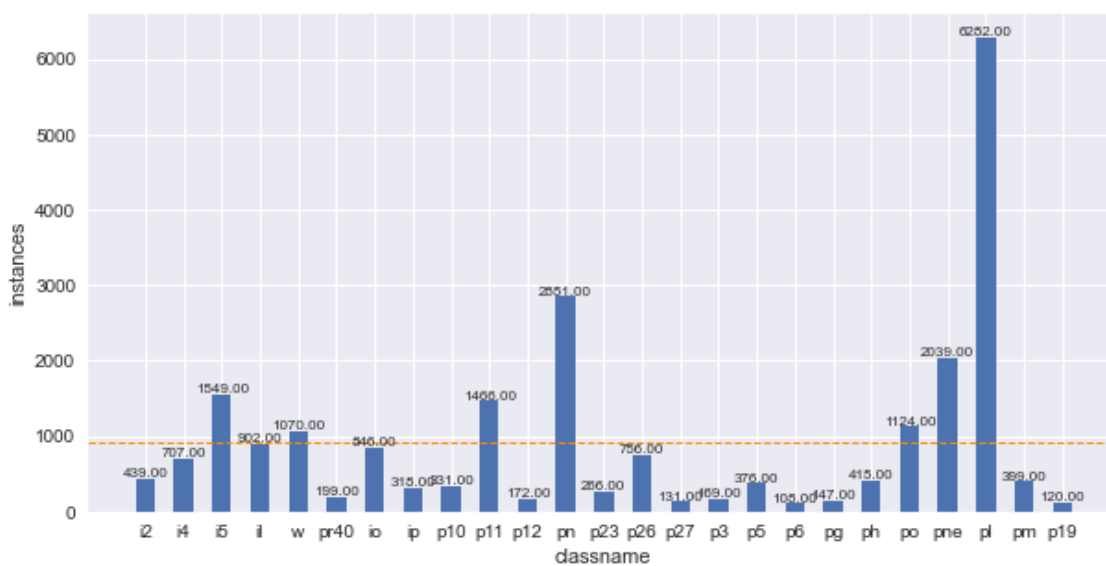

Figure 5. Version 2.0 Dataset Instances per Class

Although the merging of subdivided classes (different levels of speed limit signs) is to merge different classes together, this paper argues that this process in TSR is still equivalent to increasing the number of instances of a class in essence. Take the COCO dataset as an example, in the car class, the view angle (big difference between front view and side view), and color change in the class are much bigger than the difference between speed limit signs (only the number change in the middle of the signs). Therefore, it can be considered that the effect of the merging of subdivided classes is equivalent to increasing the number of instances of a class, and the difference of numbers on the signs will not interfere with model training. After training on the Version 2.0 dataset, the results obtained are shown in the table below. Both mAPs are significantly increased, which confirms the conjecture mentioned above in Version 1.

Table 3. Version 2.0 Dataset Training Results

| Dataset | Class | Picture | Instances | mAP@.5 | Improve | mAP@.5:.95 | Improve |
|---|---|---|---|---|---|---|---|
| Version1 | 45 | 9170 | 23182 | 0.8518 | / | 0.6546 | / |
| Version2.0 | 24 | 9170 | 23182 | 0.8816 | +3.50% | 0.6679 | +2.03% |

Based on Version 2.0, Version 2.1 using TSR dataset augmentation to augment some classes with poor training results verify TSR dataset augmentation's effectiveness. We select 150 images with 317 instances from the dataset to form "Augmentation Dataset 1". We selected six classes with poor mAP in the training results of Version 2.0 to be augmented by using TSR dataset augmentation with "Augmentation Dataset 1". The potential impact of similar backgrounds with "Augmentation Dataset 1" is ignored here.

The following figure shows the number of instances in each class of the Version 2.1 dataset after amplification. Green is Version 2.0, and blue is Version 2.1 (p19 does not exist in Version 2.0, it has been merged into the pt class. Version 2.1 restores it to a separate class and augments it. The blue part is the augmentation part.

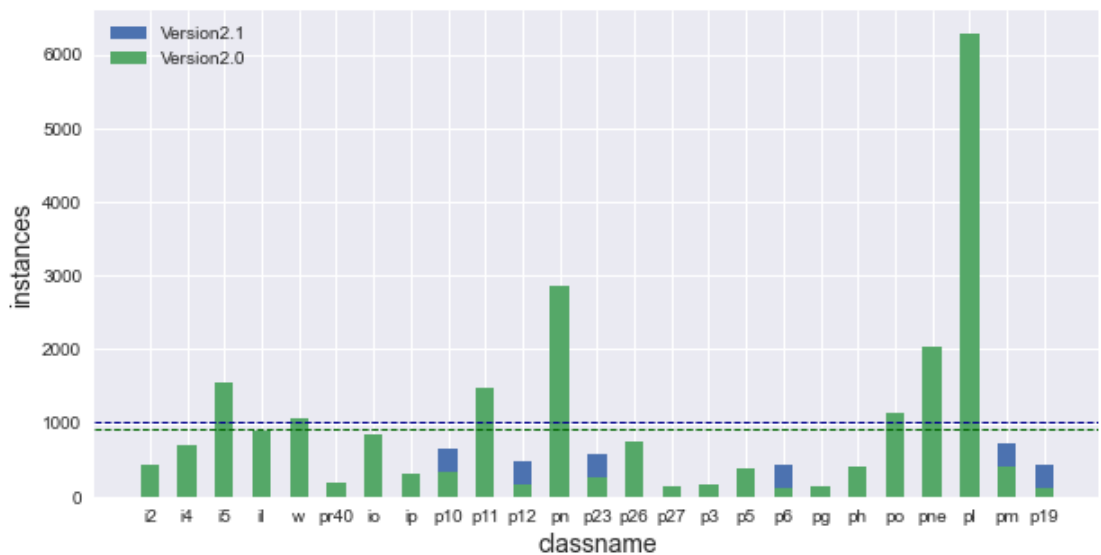

Figure 6. Version 2.1 Dataset Instances per Class

YOLOv5s Baseline model was used for training. The training results of Version 2.1 in 6 classes with augmentation were compared with Version 2.0, which were significantly better than those before augmentation. The mAP of the overall model improved slightly, but less than we expected.

Table 4. 6 Classes With Augmentation

| Class | Data Augmentation | mAP@.5 V2.0 | mAP@.5 V2.1 | Improve | mAP@.5:.95 V2.0 | mAP@.5:.95 V2.1 | Improve |
|---|---|---|---|---|---|---|---|
| all | 1902 | 0.882 | 0.887 | +0.57% | 0.668 | 0.674 | +0.90% |
| p10 | 317 | 0.773 | 0.824 | +6.60% | 0.589 | 0.622 | +5.60% |
| p12 | 317 | 0.801 | 0.858 | +7.12% | 0.599 | 0.685 | +14.35% |
| p23 | 317 | 0.868 | 0.89 | +2.53% | 0.708 | 0.716 | +1.13% |
| p6 | 317 | 0.794 | 0.859 | +8.19% | 0.598 | 0.67 | +12.04% |
| pm | 317 | 0.832 | 0.936 | +12.5% | 0.7 | 0.776 | +10.86% |
| p19 | 317 | / | 0.907 | / | / | 0.698 | / |

For the reason of low mAP improvement, the cause of the problem is found by monitoring model changes between epochs during the training process. It is found that with the increase of mAP, some signs cannot be detected. Taking 148.jpg as an example, figure A shows the effect of weights during detection in the middle training period. In the figure, all four minimum speed limit signs can be detected. While figure B shows the best weight saved in the later period of training 300 epoch, with only two speed limit signs detected.

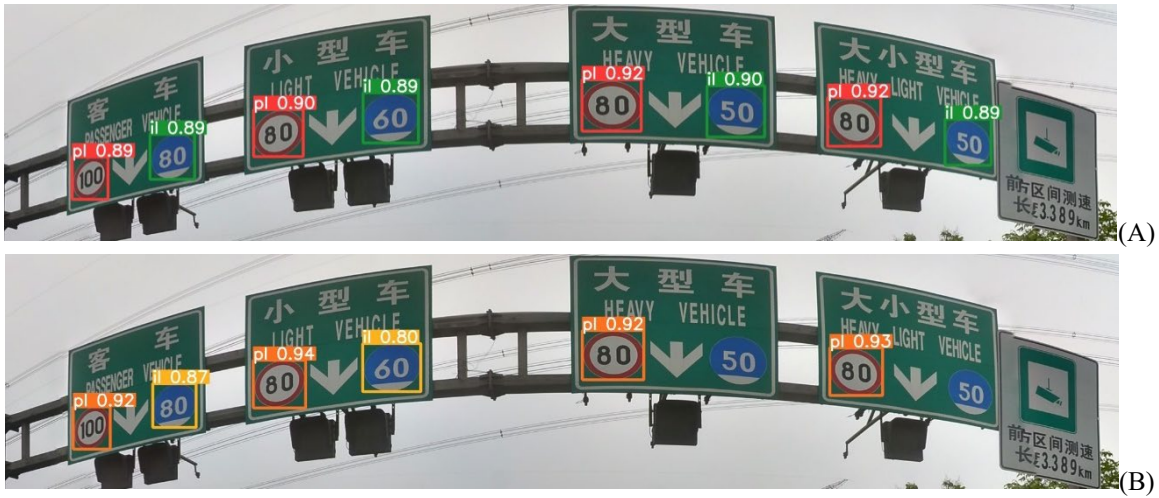

Figure 7. Detection Result. (A) weight in the middle training period; (B) best weight.

With the increase of mAP, it is very abnormal that more signs cannot be detected. A re-examination of the dataset revealed that the reason for this is that il50 (minimum speed limit 50) was not included in the Version 1 dataset because there were fewer than 100 instances. Since Version 2 is based on Version 1, il50 is also not included. Therefore, with the progress of training, il50 will be misrecognized as il in the validation set. As the model keeps improving mAP, the misrecognition will be continuously reduced, so il50 will not be recognized as il in the later stage. As for the change of il50 detection results, it only reflects the continuous optimization and error correction of the training process, and combined with the low improvement compared with Version 2.0, this paper believes that in the validation set, there are still some classes like il50 is still in false detection, which lowers the overall mAP. Since nearly 4000 instances were deleted from the original dataset, the dataset for training must be accurate, so it is necessary to reconstruct the dataset from the original dataset.

The Version 2 dataset has verified the conjecture and TSR dataset augmentation method proposed in this paper, but the improvement is lower than expected. Version 3 dataset is rebuilt from the original dataset for further validation.

### 3.3 Version 3 Dataset

Version 3.0 dataset is based on the original dataset, following the idea of Version 2, that subdivided classes are merged into one class, with a total of 28 classes.

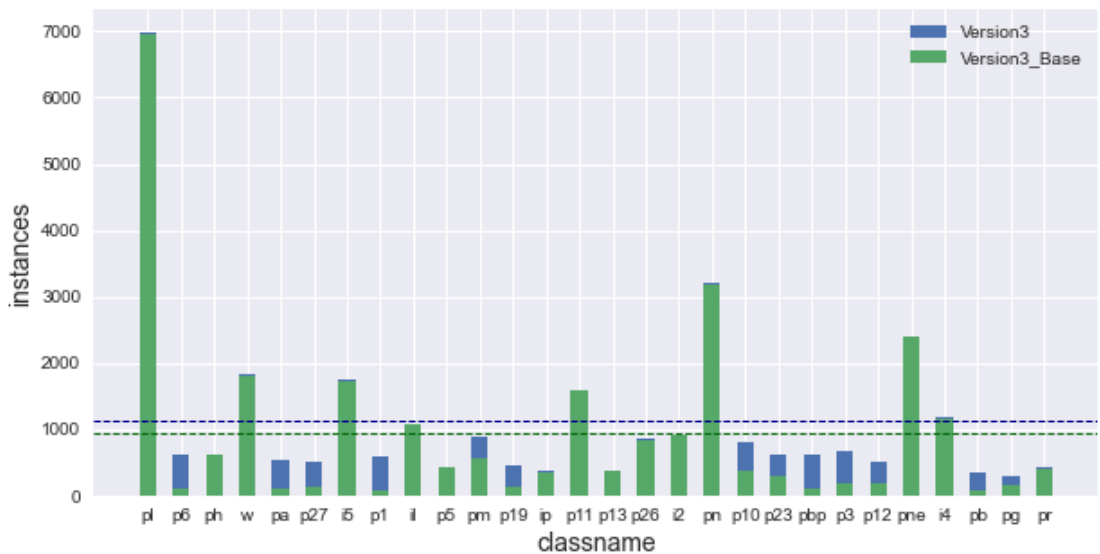

Figure 8. Version 3 Dataset Instances per Class

According to the training results of Version 2, we selected 13 classes to be augmented by TSR dataset augmentation. To avoid too many duplicate backgrounds, the number of augmentation datasets increased to 2, " Augmentation Dataset 1" containing 321 instances and " Augmentation Dataset 2" containing 436 instances.

Version 3 dataset further expands the use of TSR dataset augmentation, because the target of the previous TSR dataset augmentation is the regular round sign this time we augment the pg (Yield sign) class. Its shape is an inverted triangle, and only this kind of sign has this shape. Therefore, in order to augment this class, we need to filter out all the images containing pg and then use TSR dataset augmentation to augment. We also take advantage of traffic sign standards. As for the pb (emergency lane prohibition) class, since most of it appears on highway signs with a green background, and according to the national standards, il (minimum speed limit) signs must appear together with pl (maximum speed limit) signs and most of them appear on a green background. which is the same as the pb. So, we can filter out all the images containing il for augmentation. The augmentation of the above two classes is shown in the figure below. The Version 3 dataset augments a total of 13 classes, increasing the number of images from 9170 to 12,547, and the number of instances to 31,375.

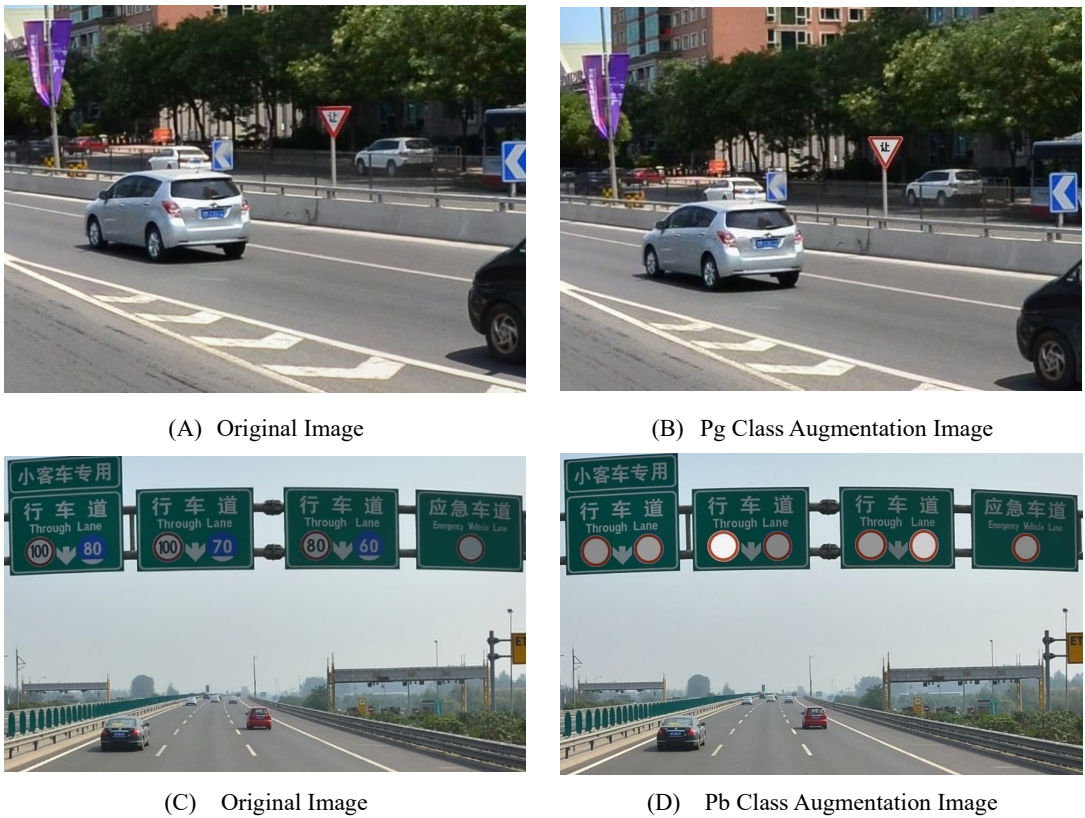

(A)　Original Image　　　　　　　　　　(B)　Pg Class Augmentation Image

(C)　Original Image　　　　　　　　　　(D)　Pb Class Augmentation Image

Figure 9. TSR Dataset Augmentation

Training on the Version 3 dataset yields the following data. It can be seen that the inclusion of all instances that need to be merged into the subdivided classes and with a bigger data augmentation makes a significant improvement over Version 2.1, even with a further increase in classes (from 25 to 28). Therefore, the above inference about the reasons for little improvement in model performance after training is correct, and it is fully confirmed that the TSR dataset augmentation method proposed in this paper is completely feasible.

**Table 5. Version 3 Dataset Training Results**

| Dataset | Class | Picture | Instances | mAP@.5 | Improve | mAP@.5:.95 | Improve |
| --- | --- | --- | --- | --- | --- | --- | --- |
| Version1 | 45 | 9170 | 23182 | 0.8518 | / | 0.6546 | / |
| Version2.0 | 24 | 9170 | 23182 | 0.8816 | +3.50% | 0.6679 | +2.03% |
| Version2.1 | 25 | 10070 | 25084 | 0.8867 | +0.57% | 0.6741 | +0.90% |
| Version3 | 28 | 12547 | 31375 | 0.9256 | +4.39% | 0.7326 | +8.68% |

In order to further verify this dataset, YOLOv5s P6 and YOLOv5n P6 models were used for training on the Version 3 dataset, and we introduced the concept of complexity of a dataset[11]. Dataset complexity can be measured by comparing the performance gap between the baseline model and the baseline-expanded complex model on the same dataset to the performance gap between another widely tested dataset, such as the COCO. If the performance gap between the complex model of the own customed dataset and the baseline model is small compared with the performance difference on the COCO dataset, then the complexity of the dataset can be roughly considered to be less than that of the COCO. By comparing the gap between YOLOv5s Baseline and two P6 models in the COCO dataset, the complexity of the dataset is determined. As shown in the following table, under the COCO dataset, the two mAP of YOLOv5s P6 and YOLOv5n P6 are significantly different, which are +17.10% and +19.64% respectively. However, under Version 3, there is no significant difference between them at mAP@.5, and Nano is even better than the Small model. At mAP@.5:.95, the difference is only +1.20%. It can be seen that Version 3 cannot show the performance gap between Small and Nano models in P6 compared with COCO. Therefore, this paper argues that the complexity of the Version 3 dataset is much lower than that of the COCO dataset. There is a lot of room for subdivision classes, that is, we can try to restore the previous subdivision classes to further validate TSR dataset augmentation. So, the Version 4 dataset will be dedicated to restoring some of the subdivision classes by using the TSR dataset augmentation proposed in this paper to perform data augmentation on the subdivision classes to ensure that there are sufficient instances of each class.

**Table 6. YOLOv5 P6 Training Result**

| Weights | Image Size | mAP@.5 | | mAP@.5:.95 | |
| --- | --- | --- | --- | --- | --- |
| | | Version 3 | COCO | Version 3 | COCO |
| YOLOv5s Baseline | 640 | 0.9256 | 0.568 | 0.7326 | 0.374 |
| YOLOv5s P6 | 1280 | 0.9719 | 0.637 | 0.7927 | 0.448 |
| YOLOv5n P6 | 1280 | 0.9726 | 0.544 | 0.7833 | 0.360 |

### 3.4 Version 4 Dataset

The Version 4 dataset contains highly similar subdivision classes of pl (maximum speed limit), and il (minimum speed limit), up to a total of 50 classes. In Version 1, the number of instances of some speed limit classes is far less than 100, making it impossible to train. This problem will be solved by using the TSR dataset augmentation proposed in this paper. In contrast to Version 3, the Version 4 dataset restores the pl and il classes to subdivision classes with data augmentation, leaving the rest of the classes unchanged.

Data augmentation in Version 4 focuses on pl and il classes. The augmentation goal is to reach a minimum of 1000 instances for the key subdivision classes (pl, il). Based on the TSR dataset augmentation proposed in this paper. Compared with the previous replacement of all marked areas in the whole image with a certain mark of a class, this time we refer to the area under the designated class's mark, which is randomly replaced with the mark in the subdivision class, and the remaining marks are replaced with their corresponding marks. This will help avoid the impact of certain categories such as pb (emergency lane ban) being restricted to specific locations. The process is shown in Figure 10 below.

Firstly, following the idea above, the subdivision classes of il were augmented. In the dataset, images with the il (minimum speed limit) are screened out, and these marks all appear on the highway speed limit signs with green backgrounds. For the subdivision classes of pl, the maximum speed limit above 80 is screened out to be augmented together with the il class (green part). For other subdivision classes of pl, images containing pl but not il were selected for augmentation, ensuring that they do not appear on the highway speed limit signs with green backgrounds (blue part). Because most of the low-speed signs cannot appear on the green speed limit signs on the highway, the impact of the potential background can be avoided. There are another thirteen classes that are augmented like Version 3. These signs belong to the regular signs in the city and appear in various locations. The augmentation is performed in two groups, using the Augmentation Datasets from Version 3 (orange and gray). pg with a special shape is still augmented like Version 3 (yellow). Other classes without color marks are not intentionally augmented and they are augmented during pl and il augmentation, which was augmented according to the original class.

**Table 7. Version 4 Dataset Augmentation Details**

| Class | Base | Version4 | Increase | Class | Base | Version4 | Increase | Class | Base | Version4 | Increase |
|---|---|---|---|---|---|---|---|---|---|---|---|
| pl80 | 904 | 1372 | 468 | pb | 76 | 683 | 607 | pn | 3176 | 7554 | 4378 |
| p6 | 116 | 770 | 654 | pg | 157 | 540 | 383 | p10 | 374 | 1079 | 705 |
| ph | 618 | 1288 | 670 | pr | 407 | 542 | 135 | p23 | 297 | 813 | 516 |
| w | 1806 | 2881 | 1075 | pl5 | 537 | 1504 | 967 | pbp | 96 | 794 | 698 |
| pa | 105 | 624 | 519 | pl10 | 39 | 1165 | 1126 | p3 | 173 | 1402 | 1229 |
| p27 | 135 | 691 | 556 | pl15 | 93 | 1185 | 1092 | p12 | 189 | 771 | 582 |
| i5 | 1734 | 3038 | 1304 | pl20 | 164 | 1289 | 1125 | pl90 | 56 | 1375 | 1319 |
| p1 | 74 | 721 | 647 | pl25 | 8 | 1091 | 1083 | pl100 | 673 | 1279 | 606 |
| il70 | 14 | 1123 | 1109 | pl30 | 640 | 1532 | 892 | pl110 | 46 | 1423 | 1377 |
| p5 | 421 | 608 | 187 | pl35 | 22 | 1100 | 1078 | pl120 | 298 | 1336 | 1038 |
| pm | 562 | 1185 | 623 | pl40 | 1413 | 1783 | 370 | il50 | 33 | 1137 | 1104 |
| p19 | 129 | 535 | 406 | pl50 | 1073 | 1495 | 422 | il60 | 489 | 1120 | 631 |
| ip | 355 | 661 | 306 | pl60 | 835 | 1224 | 389 | il80 | 297 | 1213 | 916 |
| p11 | 1582 | 4135 | 2553 | pl65 | 2 | 1052 | 1050 | il90 | 78 | 1218 | 1140 |
| p13 | 379 | 772 | 393 | pl70 | 150 | 1201 | 1051 | il100 | 132 | 1263 | 1131 |
| p26 | 840 | 2232 | 1392 | pne | 2384 | 4799 | 2415 | il110 | 21 | 1183 | 1162 |
| i2 | 901 | 1830 | 929 | i4 | 1149 | 2397 | 1248 | | | | |

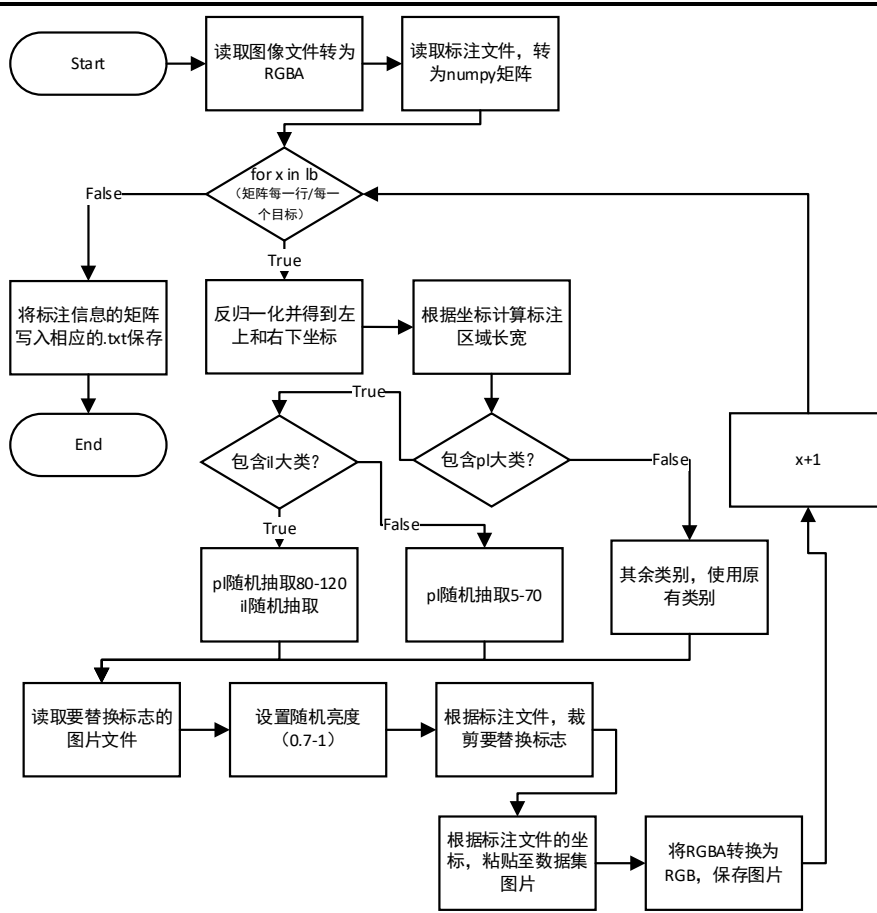

Figure 10. TSR Dataset Augmentation Process

The figure below shows the augmentation of various classes. It can be seen that the number of instances of many classes depended entirely on augmentation, such as il110, which accounted for 5533% of the data augmentation. When TSR dataset augmentation is adopted, the augmentation classes can also be adjusted dynamically. Taking each subdivision class of il as an example, the augmentation dataset which contains il pictures is used 6 times. Since the number of instances in the original dataset of each subdivision class of il is very different, for example, il60 is 489, and il110 is 21, When il60 has reached more than 1000 instances, in the subsequent augmentation, il60 is removed from the random augmentation which effectively improves the augmentation efficiency of the remaining classes and keeps the number of instances between the categories relatively average. Similarly, on the highway speed limit signs with green backgrounds, the positions of pl, il, and pb do not affect the model training. So, when the number of instances is too far away from the target 1000, the il class can be put together into the random augmentation set of the pl classes and pasted to the position of the original pl classes. This reduces training time by minimizing the amount of data needed to achieve 1000 instances per category. Through the above augmentation, the number of images in the Version 4 dataset reaches 24221, which is twice Version 3, and the number of instances is 74008, which is 2.36 times that of Version 3.

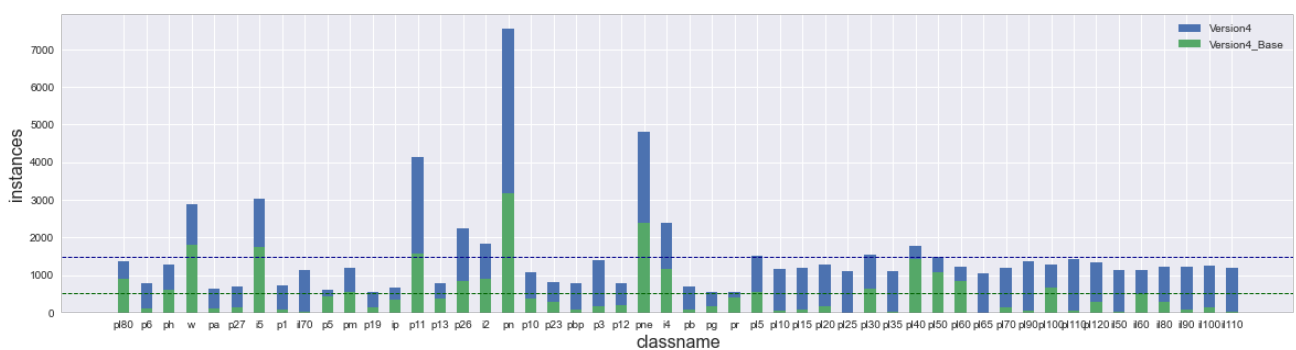

Figure 11. Version 4 Dataset Instances per Class

The following table shows the results of YOLOv5s Baseline model training on the Version 4 dataset. Compared to Version 3, there is a big drop at mAP@.5, but a significant increase at mAP@.5:.95. It is still far better than the Version 1 dataset, especially with 50 classes which are 5 more classes than Version 1. So that indicates our TSR dataset augmentation worked.

**Table 8. Version 4 Dataset Training Results**

| Dataset | Class | Picture | Instances | mAP@.5 | Improve | mAP@.5:.95 | Improve |
| --- | --- | --- | --- | --- | --- | --- | --- |
| Version 1 | 45 | 9170 | 23182 | 0.8518 | / | 0.6546 | / |
| Version 2.0 | 24 | 9170 | 23182 | 0.8816 | +3.50% | 0.6679 | +2.03% |
| Version 2.1 | 25 | 10070 | 25084 | 0.8867 | +0.57% | 0.6741 | +0.90% |
| Version 3 | 28 | 12547 | 31375 | 0.9256 | +4.39% | 0.7326 | +8.68% |
| Version 4 | 50 | 24221 | 74008 | 0.8777 | -5.18% | 0.7495 | +2.31% |

But it is important to determine whether the TSR dataset augmentation proposed in this paper is effective in large-scale applications. This can be determined by comparing the performance gap between the baseline model and the complex model training on the same dataset. And whether the training results can be kept at the same level by using more complex models. Compare the training results on Version 3 by Small and Nano models using YOLOv5 P6.

**Table 9. YOLOv5 P6 Training Result**

| Weights | Image Size | mAP@.5 | | | mAP@.5:.95 | | |
| --- | --- | --- | --- | --- | --- | --- | --- |
| | | Version 3 | Version 4 | Improve | Version 3 | Version 4 | Improve |
| YOLOv5s Baseline | 640 | 0.9256 | 0.8777 | -5.18% | 0.7326 | 0.7495 | +2.31% |
| YOLOv5s P6 | 1280 | 0.9719 | 0.9739 | +0.21% | 0.7927 | 0.8562 | +8.01% |
| YOLOv5n P6 | 1280 | 0.9726 | 0.9574 | -1.56% | 0.7833 | 0.8311 | +6.10% |

As can be seen from the above table, both YOLOv5s Baseline and YOLOv5n P6 have decreased to varying degrees on mAP@.5, while YOLOv5s P6 has increased slightly. At mAP@.5:.95, there was a significant rise in all three. Combined with the above table, it is shown that for a complex model with better performance such as YOLOv5s P6, by using the TSR dataset augmentation method proposed in this paper, the model trained on a dataset with fewer classes can outperform even with a large increase in the number of classes. Compared with the performance of the YOLOv5 P6 Small model and Nano model in Version 4 and Version 3 with the COCO dataset, the Version 4 dataset has just opened up the performance gap between P6 Small model and Nano model. However, the gap is much smaller than that between the two in the COCO dataset. Therefore, it can be inferred that the augmented dataset, when the subdivision class is increased from 28 to 50, has a limited increase in dataset complexity and still has great augmentation potential. It shows that this method is effective and has great popularizing value for data augmentation of the traffic sign recognition dataset.

## 4. CONCLUSION

By using the TSR dataset augmentation method proposed in this paper, some classes' number of instances in the original dataset is too small to be trained, such as il110 can be trained. In the TT100K dataset, there are only 21 instances of il110, and the proportion of the augmentation image is 55.3 times that of the real collected instances. It can be said that the training is completely dependent on the pictures produced by TSR dataset augmentation, and the test results can still be no different from those of other categories such as pl80 (most of its instances are from the original dataset, without data augmentation). And even when we increase the class from 45 to 50, the training results with augmentation in the Version 4 dataset are still better than the Version 1 dataset. That strongly proves that our TSR dataset augmentation is effective. It also demonstrates its ability in solving TSR training with the extremely low appearance frequency of certain classes of traffic signs in the real world. It also improves TSR dataset usability and reduces dataset collection costs so that we can collect a smaller dataset and use TSR dataset augmentation to make it trainable like a bigger dataset. The idea of TSR dataset augmentation can also be expanded the scope of applications that are not only reserved for TSR. Although this idea is only valid for the 2D object, we can at least apply it to the semantic segmentation dataset. To carry the idea further, we can replace the labeled section with semantic segmentation mask to expand its applications such as autonomous driving.

Above all, the TSR dataset augmentation is effective in solving the problem in the TSR training and its main idea has the potential to further use in various kinds of dataset augmentation.

## 5. APPENDIX

| 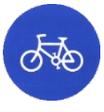 | 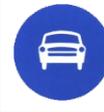 | 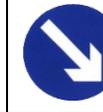 | 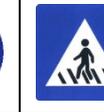 | 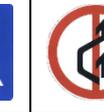 | 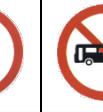 | 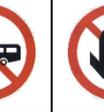 | 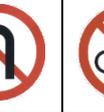 | 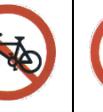 | 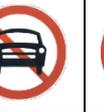 |
|---|---|---|---|---|---|---|---|---|---|
| i2 | i4 | i5 | ip | p1 | p3 | p5 | p6 | p10 | p11 |
| 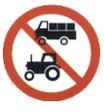 | 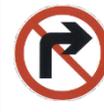 | 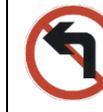 | 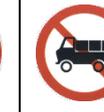 | 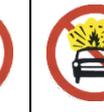 | 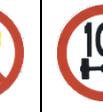 | 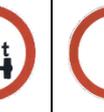 | 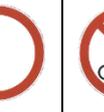 | 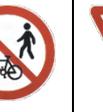 | 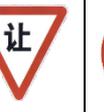 |
| p13 | p19 | p23 | p26 | p27 | pa | pb | pbp | pg | ph |
| 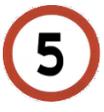 | 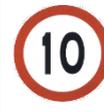 | 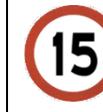 | 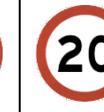 | 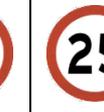 | 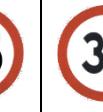 | 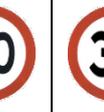 | 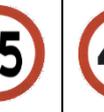 | 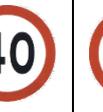 | 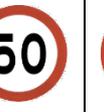 |
| pl5 | pl10 | pl15 | pl20 | pl25 | pl30 | pl35 | pl40 | pl50 | pl60 |
| pl65 | pl70 | pl80 | pl90 | pl100 | pl110 | pl120 | il50 | il60 | il70 |
| il80 | il90 | il100 | il110 | pm | pn | pne | pr | ps | w |